\documentclass[10pt,twocolumn,letterpaper]{article}

\usepackage{cvpr}              
\usepackage{makecell}
\usepackage{multirow}
\usepackage{diagbox}
\usepackage{pifont} 
\usepackage{colortbl}
\usepackage[dvipsnames]{xcolor}
\definecolor{c1}{HTML}{fae5d3}
\definecolor{c2}{HTML}{d3faf9} 
\usepackage{makecell}
\usepackage{lineno}
\definecolor{cvprblue}{rgb}{0.21,0.49,0.74}
\usepackage[pagebackref,breaklinks,colorlinks,allcolors=cvprblue]{hyperref}

\def\paperID{9675} 
\def\confName{CVPR}
\def\confYear{2025}

\title{MExD: An Expert-Infused Diffusion Model for Whole-Slide Image Classification}

\author{
    Jianwei Zhao$^{1}$, 
    Xin Li$^{2}$, 
    Fan Yang$^{2}$\thanks{Corresponding author: Fan Yang (\emph{fanyang\_uestc@hotmail.com})}, 
    Qiang Zhai$^{3}$,
    Ao Luo$^{4}$, 
    Yang Zhao$^{1}$,\\
    Hong Cheng$^{1}$, 
    and Huazhu Fu$^{5}$\\
    $^1$UESTC,
    ~~~$^2$AIQ,
    ~~~$^3$SICAU,
    ~~~$^4$SWJTU,
    ~~~$^5$IHPC, A*STAR\\
}

\begin{document}
\maketitle
\begin{abstract}
Whole Slide Image (WSI) classification poses unique challenges due to the vast image size and numerous non-informative regions, which introduce noise and cause data imbalance during feature aggregation. To address these issues, we propose MExD, an Expert-Infused Diffusion Model that combines the strengths of a Mixture-of-Experts (MoE) mechanism with a diffusion model for enhanced classification. MExD balances patch feature distribution through a novel MoE-based aggregator that selectively emphasizes relevant information, effectively filtering noise, addressing data imbalance, and extracting essential features. These features are then integrated via a diffusion-based generative process to directly yield the class distribution for the WSI. Moving beyond conventional discriminative approaches, MExD represents the first generative strategy in WSI classification, capturing fine-grained details for robust and precise results. Our MExD is validated on three widely-used benchmarks—Camelyon16, TCGA-NSCLC, and BRACS—consistently achieving state-of-the-art performance in both binary and multi-class tasks. Our code and model are available at \url{https://github.com/JWZhao-uestc/MExD}.

\end{abstract}


\section{Introduction}
\label{sec:intro}
Histological Whole Slide Image (WSI) classification is essential in digital pathology, automating diagnosis and subtyping while extracting insights from high-resolution scans to support prognosis and treatment planning~\cite{bejnordi2017diagnostic,madabhushi2009digital,cornish2012whole}. WSIs, with sizes ranging from 100 million to 10 billion pixels, are too large for conventional deep learning techniques, which were originally designed for much smaller natural or medical images. To tackle this, the ``\emph{decompose and aggregate}'' strategy has proven effective, breaking down WSIs into manageable patches to extract features and then combining them into a final representation for classification. A promising application of this strategy is Multiple Instance Learning (MIL), which treats each WSI as a \emph{bag} containing smaller image patches. By analyzing each patch individually and then aggregating the results, MIL enables efficient slide-level classification with a single label, effectively addressing the challenges of WSI analysis~\cite{amores2013multiple,dietterich1997solving}. 

However, despite its effectiveness, this strategy faces several challenges. A significant issue is data imbalance: within a \emph{bag}, \emph{positive} instances (\emph{e.g.}, those containing cancerous cells) are often vastly outnumbered by \emph{negative} ones. This imbalance can bias the model toward the majority class, leading it to overlook or misclassify important minority instances and resulting in poor sensitivity for detecting rare but clinically significant features. Although clustering~\cite{zhang2022dtfd,li2023task} and attention mechanisms~\cite{lu2021data,li2021dual} have been introduced to tackle this issue, these methods may still not fully mitigate the bias toward the majority class. This is especially true when \emph{positive} instances are extremely scarce; under such circumstances, these mechanisms may be less effective because they typically operate within a unified model that processes all patches collectively, lacking dedicated components focused on minority classes.

\begin{figure}[pt]
	\begin{center}
		\includegraphics[width=0.99\linewidth]{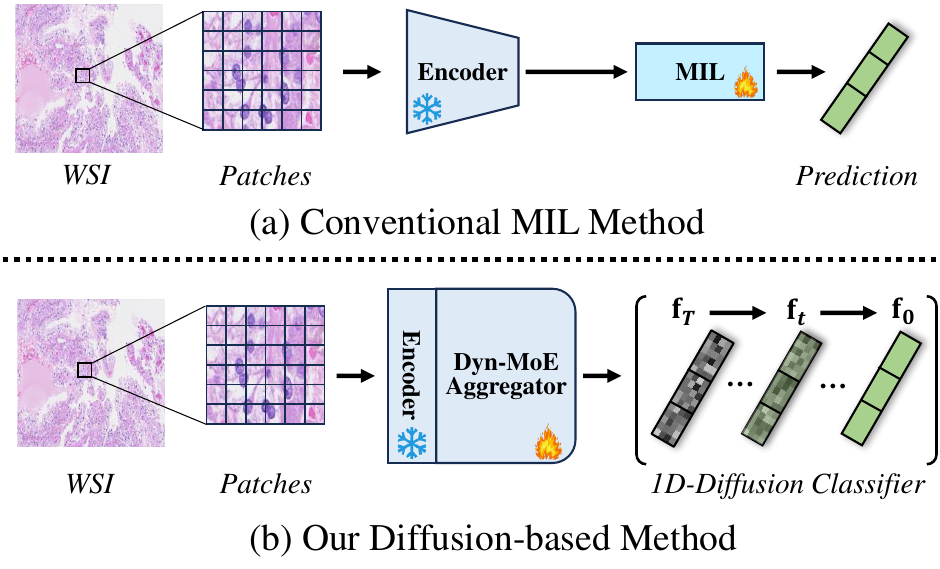}
	\end{center}
		\vspace{-10pt}
	\caption{{\bf Idea Illustration}. (a) Unlike the conventional MIL-based discriminative approaches for WSI processing, (b) we employ the Dyn-MoE to extract/mine effective conditional information from WSI, employing a generative framework to infer classification results.}
	\label{fig:1}
			\vspace{-15pt}
\end{figure}

Another challenge lies in the current shortcomings of feature integration methods. Aggregating the extracted features from individual patches into a comprehensive slide-level representation is non-trivial. Standard aggregation methods—such as simple averaging or max pooling—may be inadequate for capturing the intricate interrelationships and spatial dependencies among patches. Some advanced techniques, including reinforcement learning to guide feature integration~\cite{zheng2024dynamic} and the use of dynamic graph models~\cite{li2024dynamic}, have been employed to better learn integrated features. However, these methods are still limited by the influence of noise and irrelevant patches, as they fundamentally rely on discriminative feature integration paradigms that are inherently susceptible to such interference.

To address these challenges, we introduce the Expert-Infused Diffusion Model (MExD), a novel paradigm that integrates the Mixture-of-Experts (MoE) mechanism with a diffusion-based conditional generative model, as illustrated in Fig.~\ref{fig:1}. A key component of MExD is the Dynamic Mixture-of-Experts (Dyn-MoE) aggregator, which intelligently routes instances to specialized experts based on their relevance and characteristics. This dynamic routing mechanism ensures focused attention on minority classes, enhancing sensitivity to rare but clinically significant instances and reducing bias toward the majority class. In essence, Dyn-MoE extracts essential conditional information—drawing from both expert insights and prior prediction—to effectively guide subsequent operations.



Moreover, we introduce a novel generative-based learning and inference approach within MExD for WSI classification. Our Diffusion Classifier (Diff-C) leverages the expert and prior knowledge extracted by Dyn-MoE through a customized diffusion process, supported by a multi-layer perceptron (MLP) network. This integration of expert-driven insights with the generative capabilities of the diffusion model enables the Diff-C to directly predict accurate and reliable slide-level classifications. Additionally, the diffusion-based generative approach excels at mitigating noise and effectively capturing the relationships within the underlying data distribution, thereby enhancing the robustness of the model's learning and inference. This innovative strategy not only captures intricate interdependencies among patches but also ensures that the final predictions are coherent and meaningful, significantly improving classification performance and reliability in WSI analysis. Overall, the {\bf contributions} of this work can be summarized as:
\begin{itemize}
	\item{\bf  A new paradigm for WSI classification.} MExD is the first to apply generative techniques for WSI classification, offering a novel solution to data imbalance, noise reduction, and feature integration, and laying the foundation for future research.
 
	\item{\bf A novel diffusion model fused with the MoE approach.} We introduce a novel diffusion model seamlessly integrated with Mixture-of-Experts (MoE), which, unlike standard diffusion models, incorporates experts focused on specific instances, thereby more effectively addressing the inherent challenges of WSI analysis.
 
	\item{\bf State-of-the-art results on widely-used benchmarks.} MExD demonstrates exceptional performance in accurately classifying WSI, setting new standards on prominent benchmarks such as Camelyon16, TCGA-NSCLC, and BRACS.
\end{itemize}



\section{Related Work}
\label{sec:related}
\noindent \textbf{Whole Slide Image Classification.}
MIL-based models play a key role in WSI classification, with most approaches employing pooling~\cite{campanella2019clinical,lu2021data} and attention mechanisms~\cite{ilse2018attention,li2021dual,shao2021transmil} to identify predictive features. However, these methods often overlook inter-instance correlations, limiting their effectiveness. Recent techniques have begun to address this by clustering or randomly grouping instances into pseudo-bags or labels~\cite{qu2022dgmil,zhang2022dtfd,zhu2022murcl,chen2023rankmix,li2023task,shao2023lnpl}, enhancing the model’s inference capabilities. Others have incorporated multi-resolution WSIs~\cite{thandiackal2022differentiable,li2021dual,chen2022scaling}, reinforcement learning~\cite{zheng2024dynamic}, causal learning~\cite{chen2024camil,zhang2022multi}, VLM~\cite{li2024generalizable,qu2024rise}, and graph models~\cite{li2024dynamic,hou2022h,guan2022node,chan2023histopathology} to further boost accuracy. However, as discriminative models, they still encounter challenges such as data imbalance and noise sensitivity. To overcome these limitations, we propose MExD, a generative approach that addresses these issues, offering enhanced robustness in WSI analysis.

\noindent \textbf{Mixture of Experts (MoE).} 
Modern MoE methods, inspired by~\cite{shazeer2017outrageously}, introduced the Sparsely-Gated Mixture-of-Experts layer as a replacement for the feed-forward network in language models, allowing for efficient and adaptive computations. In computer vision and medical imaging, MoE has been successfully applied to tasks such as image fusion~\cite{zhu2024task}, scene decomposition~\cite{zhenxing2022switch}, and histopathology analysis~\cite{li2024m4}. Building on this concept, we develop an embedded module called Dyn-MoE, specifically designed to extract critical WSI information as conditional input to enhance inference in our framework.

\noindent \textbf{Diffusion Models.} 
Diffusion models are powerful generative methods that iteratively refine data through noise-based transformations, creating realistic outputs from random noise~\cite{ho2020denoising,yang2023diffusion,zhao2024focusdiffuser,luo2024flowdiffuser}. In medical imaging, they have shown promise in tasks such as image generation~\cite{puglisi2024enhancing,zhu2024generative,zhan2024medm2g,chen2024towards}, denoising~\cite{xiang2023ddm}, and segmentation~\cite{kim2022diffusion,zbinden2303stochastic}. Despite these successes, diffusion models have not yet been applied to WSI classification, which is particularly challenging due to the complexity and massive scale of WSIs. In this work, we introduce the Expert-Infused Diffusion Model, the first diffusion-based approach specifically designed for WSI classification. Our model incorporates a Mixture-of-Experts (MoE) mechanism within the diffusion framework to address the unique characteristics of WSIs. To the best of our knowledge, this is the first approach to integrate MoE into diffusion-based learning and inference, opening new possibilities for WSI analysis and beyond.

\begin{figure*}[pt!]
	\begin{center}
		\includegraphics[width=0.99\linewidth]{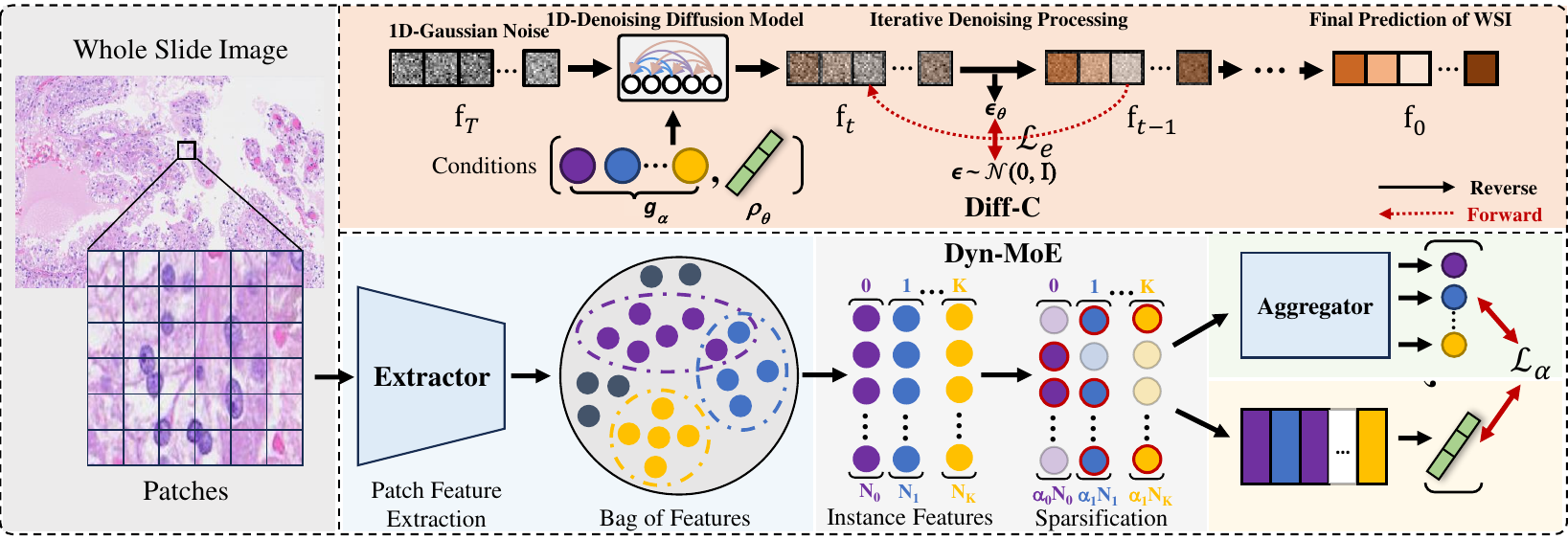}
	\end{center}
		\vspace{-10pt}
\caption{{\bf Overview of MExD}. Our framework leverages a generative classification diffusion model for enhanced reliability and generalizability, where $\boldsymbol{g}_{\alpha}$ and $\boldsymbol{\rho}_{\theta}$ encode expert insights and prior knowledge, conditioning Diff-C (see Sec.~\ref{sec:mexd} for details).}
	\label{fig:2}
			\vspace{-10pt}
\end{figure*}

\section{Method}
\label{sec:method}
\subsection{Preliminaries}
\noindent \textbf{Motivation.} 
Despite extensive exploration, discriminative models still fall short in WSI classification due to challenges like massive image sizes, noise, and data imbalance from non-informative regions. These limitations highlight the need for a paradigm shift, prompting us to explore a generative approach that better captures the complexity and variability inherent in WSI data.


\noindent \textbf{Problem Reformulation.} 
Whole slide image (WSI) classification is typically approached as a Multiple Instance Learning (MIL) problem due to the gigapixel resolution, which makes direct input into neural networks impractical. Current methods are predominantly \emph{discriminative}, where a WSI ($\boldsymbol{X}$) is divided into a patch-wise \emph{bag} $\boldsymbol{\breve{B}}$ of $N$ patches $\{ \boldsymbol{x}_i \}_{i=1}^N$, with only a bag-level label $\boldsymbol{Y}$. For binary classification, $\boldsymbol{Y} = 1$ indicates at least one \emph{positive} instance, while $\boldsymbol{Y} = 0$ indicates none. In multi-label tasks, $\boldsymbol{Y}$ corresponds to the specific cancer sub-type, with each WSI containing only one type. The \emph{discriminative} MIL workflow generally includes three steps: 1) \textbf{Patch Embedding}: Extractor $\boldsymbol{f(\cdot)}$ transforms patches $\{\boldsymbol{x}_i\}_{i=1}^N$ into embeddings $\{ \boldsymbol{b}_i \}_{i=1}^N$, forming instance-wise \emph{bag} $\boldsymbol{B}$;  
2) \textbf{Feature Aggregation}: Aggregator $\boldsymbol{g(\cdot)}$ combines embeddings;
3) \textbf{WSI Classification}: Classifier $\boldsymbol{c(\cdot)}$ predicts the label $\boldsymbol{\hat{Y}}$: $\hat{Y}=\boldsymbol{c}(\boldsymbol{g}(\{\boldsymbol{f}(x_i)\}_{i=1}^N))$. In some methods, the classifier $\boldsymbol{c(\cdot)}$ is incorporated into $\boldsymbol{g(\cdot)}$.

In contrast to conventional approaches, we reformulate this as a conditional \emph{generative} task. We retain the \textbf{Patch Embedding} step, similar to \emph{discriminative} methods, to extract features from patches, while introducing new techniques for \textbf{Feature Aggregation} and \textbf{WSI Classification}. To accomplish this, we design a Dynamic Mixture-of-Experts (Dyn-MoE) aggregator, $\boldsymbol{g_{moe}(\cdot)}$, leveraging the MoE mechanism to sparsify instance embeddings and capture conditional information from $\boldsymbol{B}$, including expert insights and prior prediction. Using this enriched information, we construct a Diffusion Classifier (Diff-C), $\boldsymbol{c_{diff}(\cdot)}$, to directly generate accurate and reliable slide-level classifications. Formally, this process is expressed as $\mathbf{f} = \mathbb{P}_{\Theta}(\mathbf{f}_n|\boldsymbol{B})$, following generative training principles. The parameter set $\Theta$ includes trained parameters from both $\boldsymbol{g_{moe}(\cdot)}$ and $\boldsymbol{c_{diff}(\cdot)}$, with instance embeddings extracted offline by $\boldsymbol{f(\cdot)}$. Our {MExD} represents a significant advancement in leveraging the generative power of diffusion models for WSI classification, moving beyond the limitations of conventional discriminative models.


\subsection{Expert-Infused Diffusion Model}
\label{sec:mexd}
\noindent {\bf{Overview.}} Fig.~\ref{fig:2} presents an overview of our MExD, which integrates a Patch Feature Extractor, a Dyn-MoE Aggregator leveraging the Mixture-of-Experts mechanism, and a Diffusion Classifier (Diff-C) built on the principles of the Denoising Diffusion Probabilistic Model:
\begin{itemize}
    \item \noindent {\bf{Patch Feature Extractor.}} Given a patch-wise \emph{bag} $\boldsymbol{\breve{B}}=\{\boldsymbol{x}_i\}^{N}_{i=1}$, the Patch Feature Extractor, similar to conventional discriminative methods, encodes each patch to generate high-order instance features $\{\boldsymbol{b}_i\}^{N}_{i=1}$, which are then fed into the aggregation stage.
    
    
    \item \noindent {\bf{Dyn-MoE Aggregator.}} The Dyn-MoE Aggregator receives $\{\boldsymbol{b}_i\}^{N}_{i=1}$ as input and dynamically samples the most representative instances using a tailored MoE mechanism, yielding a sparse set of transformed instances $\{\boldsymbol{l}_i\}^{M}_{i=1}$, where $M<\frac{N}{2}$. It then learns the concatenation $[\{\boldsymbol{l}_i\}^{M}_{i=1}, \boldsymbol{d}]$ to produce prior prediction $\boldsymbol{\rho}_{\theta}$, which captures the relationship between the WSI ($\boldsymbol{X}$) and the bag-level label ($\boldsymbol{Y}$), with $\boldsymbol{d}$ as a learnable class embedding. Meanwhile, each expert sub-branch generates a class-centric latent feature $\boldsymbol{e}$ and confidence $\boldsymbol{c}$, forming the expert insights set $\boldsymbol{g}_{\alpha} = \{(\boldsymbol{e}_i, \boldsymbol{c}_i)\}^{K}_{i=0}$.
    
    
    \item \noindent {\bf{Diffusion Classifier.}} Our Diff-C utilizes a modified forward diffusion process, incorporating prior prediction $\boldsymbol{\rho}_{\theta}$ as a predicted conditional expectation, along with a multi-layer perceptron (MLP) serving as a noise prediction network conditioned on the expert insights set $\boldsymbol{g}_{\alpha}$, to directly predict the final result $\boldsymbol{\hat{Y}}$.
    
\end{itemize}
These components work together to form MExD, an innovative diffusion-based model for WSI classification. By leveraging the MoE mechanism, it filters out vast numbers of \emph{negative} instances and directly generates bag-level predictions through a diffusion process. Further details are provided in the following sections.


\noindent {\bf{Patch Feature Extractor (PFE).}}
Our Patch Feature Extractor, $\boldsymbol{f}_{\mathbb{PFE}}$, extracts semantic features from patches $\{\boldsymbol{x}_i \}^{N}_{i=1} \in \mathbb{R}^{h\times w \times 3}$ within a patch-wise \emph{bag} $\boldsymbol{\breve{B}}$, producing an instance-wise \emph{bag} $\boldsymbol{B}$ of embeddings $\{\boldsymbol{b}_i \}^{N}_{i=1} \in \mathbb{R}^{1\times C}$, while maintaining high-level representations. To demonstrate {MExD}'s plug-and-play flexibility, we employ two variants of $\boldsymbol{f}_{\mathbb{PFE}}$: ViT pre-trained with MoCo V3~\cite{chen2021empirical} and CTransPath pre-trained with SRCL~\cite{wang2020visual}, following IBMIL~\cite{lin2023IBMIL}.


\noindent {\bf{Dyn-MoE Aggregator.}} 
A gigapixel whole slide image yields a \emph{bag} of thousands of instance features, many of which are redundant, hindering classification if integrated directly. To address this, we designed the Dyn-MoE aggregator, which leverages the sparse Mixture-of-Experts (MoE) mechanism to select only the most discriminative instances, assigning dedicated experts for each \emph{positive} subtype and one for \emph{negative} instances.


\begin{figure}[pt]
	\begin{center}
		\includegraphics[width=0.99\linewidth]{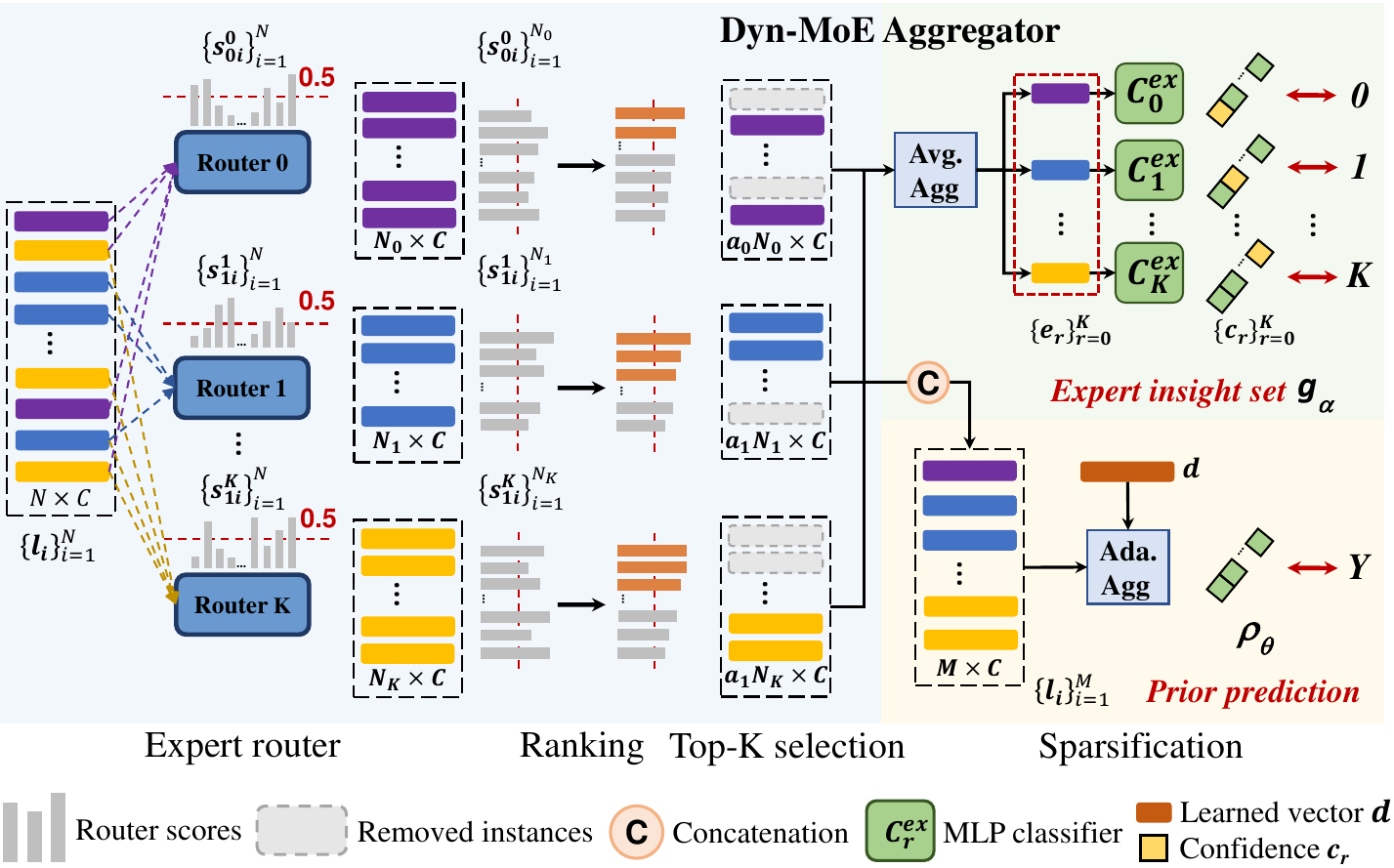}
	\end{center}
		\vspace{-10pt}
	\caption{{\bf Details of Dyn-MoE Aggregator}. Our Dyn-MoE includes K \emph{positive} experts and 1 \emph{negative} expert, producing expert insights set ($\boldsymbol{g}_{\alpha}$) and prior prediction ($\boldsymbol{\rho}_{\theta}$) to work alongside Diff-C. `Avg.' and `Ada.' indicate mean-pooling and Adapter. \emph{Zoom in for details.}}
	\label{fig:3}
		\vspace{-15pt}
\end{figure}

As shown in Fig.~\ref{fig:3}, the instance \emph{bag} $\{\boldsymbol{b}_i \}^{N}_{i=1}$ from the patch feature extractor is passed through an Adapter, consisting of two transformer layers (TPT) and a convolution block (PPEG) from TransMIL~\cite{shao2021transmil}, to produce instance features $\{\boldsymbol{l}_i\}^{N}_{i=1}$ with global dependencies. This can be formulated as:
\begin{equation}
  \{\boldsymbol{l}_i\}^{N}_{i=1} = Ada(\{\boldsymbol{b}_i\}^{N}_{i=1}).
  \label{eq:adapter}
\end{equation}

Our Dyn-MoE then applies sparsity processing to $\{\boldsymbol{l}_i\}^{N}_{i=1}$. For tasks with $K$ \emph{positive} sub-types and 1 \emph{negative}, {MExD} is equipped with $K+1$ experts. Each expert has a router network $\mathrm{R}_r$—a single-layer 2-class MLP with softmax—to sample instances. At the $r$-th routing, each instance $\boldsymbol{l}_i$ generates two scores, $\{s_{0i}, s_{1i}\}$. For the \emph{negative} expert (expert 0), the instance is routed to its sub-branch when the maximum score index is 0; for \emph{positive} experts (1 to $K$), it is routed when the index is 1. After processing by all $K+1$ routers, we obtain $K+1$ subsets, $\{ \{\boldsymbol{l}_i\}^{N_0}_{i=1}, \{\boldsymbol{l}_i\}^{N_1}_{i=1}, \cdots, \{\boldsymbol{l}_i\}^{N_{K}}_{i=1} \}$, and corresponding router scores $\{ \{\boldsymbol{s}^{0}_{0i}\}^{N_0}_{i=1}, \{\boldsymbol{s}^{1}_{1i}\}^{N_1}_{i=1}, \cdots, \{\boldsymbol{s}^{K}_{1i}\}^{N_{K}}_{i=1} \}$. To retain the most representative instances, each subset keeps only the instances with the `top-k' router scores. The routing process can be formalized as:
\begin{equation}
\begin{gathered}
  \{\boldsymbol{l}_j\}^{N_r}_{j=1} = \{\boldsymbol{l}~|~\mathrm{Argmax}(\mathrm{R}_r(\boldsymbol{l}))=\boldsymbol{\eta}, \boldsymbol{l} \in \{\boldsymbol{l}_i\}^{N}_{i=1} \}, \\
  \{\boldsymbol{l}_j\}^{{\alpha}{N_r}}_{j=1} = \{\boldsymbol{l}_i\in \{\boldsymbol{l}_j\}^{N_r}_{j=1}~|~i \in \mathrm{ArgTopK}(\{\boldsymbol{s}^{r}_{\boldsymbol{\eta}j}\}^{N_r}_{j=1})\}, \\
  \boldsymbol{\eta} = 0 \text{ if } r = 0,\; \text{ else }\boldsymbol{\eta} = 1,
  \label{eq:topk}
\end{gathered}
\end{equation}
where $\alpha$ controls the sampling ratio, and `$\mathrm{ArgTopk}$' selects indexes of instances with `top-k' router scores, where $\mathrm{k} = \alpha N_r$. After routing, each expert applies mean pooling $\boldsymbol{\sigma(\cdot)}$ to its sampled instances to produce a class-centric latent feature $\boldsymbol{e}$. Each expert then uses a $(K+1)$-class classifier $\boldsymbol{C}^{ex}(\cdot)$, consisting of a linear layer with softmax, which takes $\boldsymbol{e}$ as input and outputs class predictions $\boldsymbol{y}^{ex} \in \mathbb{R}^{1 \times (K+1)}$, yielding an expert confidence score $\boldsymbol{c}$.
\begin{equation}
\begin{gathered}
  \boldsymbol{e}_r = \boldsymbol{\sigma} (\{\boldsymbol{l}_j\}^{{\alpha}{N_r}}_{j=1}),~
  \boldsymbol{y}^{ex}_r = \boldsymbol{C}^{ex}_{r}{(\boldsymbol{e}_r)},~\boldsymbol{c}_r = \boldsymbol{y}^{ex}_r[\mathrm{:}r].
  \label{eq:confidence}
\end{gathered}
\end{equation}

We combine all subsets after the `top-k' process to obtain the final sparsified instance \emph{bag} \emph{s}-$\boldsymbol{B}=\{\boldsymbol{l}_i\}_{i=1}^{M}$. A learnable class embedding $\boldsymbol{d}$ is concatenated with \emph{s}-$\boldsymbol{B}$, and this combined representation is passed through an Adapter and a linear classifier $\boldsymbol{C}(\cdot)$ to generate the class prediction $\boldsymbol{\rho}_{\theta}$, serving as auxiliary input for the diffusion-based classifier. This process is defined as follows, where [,] denotes concatenation:
\begin{equation}
  \boldsymbol{\rho}_{\theta} = \boldsymbol{C}(Ada([\{\boldsymbol{l}_i\}_{i=1}^{M}), \boldsymbol{d}]).
  \label{eq:adapter_1}
\end{equation}
\noindent {\bf{Diffusion Classifier (Diff-C).}} 
Diffusion Models (DMs) have achieved notable success in 2-D image generation, but their application to classification tasks is limited. A standard DM consists of forward diffusion (FD) and reverse denoising (RD), mainly designed for 2-D signals. {Diff-C}, however, reformulates classification as a 1-D signal generation task. Specifically, we encode the ground truth $\boldsymbol{Y}$ as a one-hot label $\boldsymbol{\mathrm{f}} \in \mathbb{R}^{1 \times (K+1)}$ in a continuous 1-D space as the initial FD state. We further adapt the diffusion process, guided by prior prediction $\boldsymbol{\rho}_{\theta}$ and expert insights set $\boldsymbol{g}_{\alpha}$, to accomplish WSI classification.


\noindent {\bf{(a) \emph{Forward Diffusion with $\boldsymbol{\rho}_{\theta}$}}.} 
Forward diffusion generates a series of noised samples $\boldsymbol{y}_t$ at timestamps $\boldsymbol{t}$ from the initial signal $\boldsymbol{y}_0$, following a Markov process as shown in the first line of Eq.~\ref{eq:FD_ori}. As $\boldsymbol{t}$ increases from 1 to $T$, $\boldsymbol{y}_t$ can also be derived directly from $\boldsymbol{y}_0$ using the second line of Eq.~\ref{eq:FD_ori}:
\begin{equation}
\begin{gathered}
  q(\boldsymbol{y}_t|\boldsymbol{y}_{t-1}) = \mathcal{N}(\boldsymbol{y}_t; {\tilde{\beta}}_{t}\boldsymbol{y}_{t-1},~{\beta}_{t}\boldsymbol{\mathrm{I}}), \\
  q(\boldsymbol{y}_t|\boldsymbol{y}_{0}) = \mathcal{N}(\boldsymbol{y}_t; \sqrt{\overline{\alpha}_{t}}\boldsymbol{y}_{0},~(1-\overline{\alpha}_{t})\boldsymbol{\mathrm{I}}).
  \label{eq:FD_ori}
\end{gathered}
\end{equation}

Here, ${\beta}_{t}$ is the variance of the Gaussian noise at each step $\boldsymbol{t}$, with ${\tilde{\beta}}_{t} = \sqrt{1 - {\beta}_{t}}$, $\overline{\alpha}_{t} = \prod_{j=1}^{t} {\alpha}_{j}$, and ${\alpha}_{j} = 1 - {\beta}_{j}$. As $\overline{\alpha}_{T} \rightarrow 0$ at the maximum timestamp $T$, the endpoint of FD, $\boldsymbol{y}_T$ becomes pure Gaussian noise. Inspired by CARD~\cite{han2022card}, we assume the one-hot encoded label $\boldsymbol{\mathrm{f}}_0$ with $\boldsymbol{\rho}_{\theta}$ as the predicted conditional expectation, infusing it into FD to specify conditional distributions in Eq.~\ref{eq:FD_new} (top), allowing closed-form sampling at any timestamp $\boldsymbol{t}$ in Eq.~\ref{eq:FD_new} (bottom):
\begin{equation}
\begin{gathered}
  q(\boldsymbol{\mathbf{f}}_t|\boldsymbol{\mathbf{f}}_{t-1}, \boldsymbol{\rho}_{\theta}) = \mathcal{N}(\boldsymbol{\mathbf{f}}_t; {\tilde{\beta}}_{t}\boldsymbol{\mathbf{f}}_{t-1}+(1-{\tilde{\beta}}_{t})\boldsymbol{\rho}_{\theta},~{\beta}_{t}\mathrm{\boldsymbol{\mathrm{I}}}), \\
  q(\boldsymbol{\mathbf{f}}_t|\boldsymbol{\mathbf{f}}_{0}, \boldsymbol{\rho}_{\theta})=\mathcal{N}(\boldsymbol{\mathbf{f}}_t; \sqrt{{\overline{\alpha}}_{t}}\boldsymbol{\mathbf{f}}_{0}+(1\mathrm{-}\sqrt{{\overline{\alpha}}_{t}})\boldsymbol{\rho}_{\theta},{(1\mathrm{-}\overline{\alpha}}_{t})\mathrm{\boldsymbol{\mathrm{I}}}).
  \label{eq:FD_new}
\end{gathered}
\end{equation}

Intuitively, in {MExD}, the endpoint of FD is the prior prediction $\boldsymbol{\rho}_{\theta}$.

\noindent {\bf{(b) \emph{Reverse Denoising with $\boldsymbol{g}_{\alpha}$}}.}
Diff-C uses a denoising network $\mathcal{D}$, implemented as a three-layer MLP, to predict the added noise $\boldsymbol{\epsilon}_t$. It receives expert insights set $\boldsymbol{g}_{\alpha}$ and an initial noise input $\boldsymbol{\mathrm{f}}_T \sim \mathcal{N}(\boldsymbol{\rho}_{\theta}, \boldsymbol{\mathrm{I}})$. First, we compute a weighted average of all $\boldsymbol{e}$ values in $\boldsymbol{g}_{\alpha}$ using their corresponding confidences $\boldsymbol{c}$, which is fed into an encoder with three linear layers to produce the conditioned latent $\boldsymbol{Z} \in \mathbb{R}^{1 \times C}$. The denoising network iteratively estimates the intermediate noise ${\epsilon}_{\theta}$, refining $\boldsymbol{\mathrm{f}}_t$ to $\boldsymbol{\mathrm{f}}_{t-1}$ following the FD posterior from CARD~\cite{han2022card}, until the reconstructed output $\boldsymbol{\mathrm{f}}^{'}_0$ is obtained:
\begin{equation}
  q(\boldsymbol{\mathbf{f}}_{t-1}|\boldsymbol{\mathbf{f}}_{t}, \boldsymbol{\mathbf{f}}_{0}, \boldsymbol{\rho}_{\theta}) = \mathcal{N}(\boldsymbol{\mathbf{f}}_{t-1}; {\boldsymbol{\mu}}(\boldsymbol{\mathbf{f}}_{t}, \boldsymbol{\mathbf{f}}_{0}, \boldsymbol{\rho}_{\theta}), \hat{\beta}_t\boldsymbol{\mathrm{I}}).
  \label{eq:FD_poster}
\end{equation}
Based on Eq.~\ref{eq:FD_poster}, the entire process can be mathematically defined as follows:
\begin{equation}
\begin{gathered}
   \boldsymbol{Z} = \sum\limits_{r=0}^{k}(\boldsymbol{c}_r \cdot \boldsymbol{e}_r), \quad
   \boldsymbol{\epsilon}_{\theta} = \mathcal{D}(\boldsymbol{Z}, \boldsymbol{\mathrm{f}}_t, \boldsymbol{\rho}_{\theta}, \boldsymbol{t}), \\
   \boldsymbol{\hat{\mathrm{f}}}_0 = \frac{1}{\sqrt{\overline{\alpha}_t}}(\boldsymbol{\mathrm{f}}_t-(1\mathrm{-}\sqrt{{\overline{\alpha}}_{t}})\boldsymbol{\rho}_{\theta}-(1\mathrm{-}\sqrt{{\overline{\alpha}}_{t}})\boldsymbol{\epsilon}_{\theta}), \\
   \boldsymbol{{\mathrm{f}}}_{t-1} = 
   {\gamma}_0\boldsymbol{\hat{\mathrm{f}}}_0 + {\gamma}_1\boldsymbol{{\mathrm{f}}}_{t} + 
   {\gamma}_2\boldsymbol{\rho}_{\theta}+
   {\gamma}_3\boldsymbol{z}.
  \label{eq:rd}
\end{gathered}
\end{equation}
Here, $\boldsymbol{z}$ is a standard Gaussian noise. ${\gamma}_0$, ${\gamma}_1$, ${\gamma}_2$ and ${\gamma}_3$ are associated with parameters $\alpha$ and $\beta$\footnote{Derivation details are provided in the {\bf Supplementary Materials}.}. In the context of the reverse denoising process, as formalized in Eq.~\ref{eq:rd}, the Diff-C progressively refines the noisy prediction through an iterative sequence, \emph{i.e.}, $\boldsymbol{\mathrm{f}}_T \rightarrow \boldsymbol{\mathrm{f}}_{T-\Delta} \rightarrow \cdots \boldsymbol{\mathrm{f}}_0$. 

\noindent {\bf{Training Objective.}} MExD is trained in two stages to ensure stable convergence. First, we train the Dyn-MoE as an auxiliary model using a joint loss $\mathcal{L}_{a}$, which includes a cross-entropy loss $\Phi$ to supervise the prior prediction $\boldsymbol{\rho}_{\theta}$ and an MoE loss to regulate the expert sub-branches. Specifically, the MoE loss is the sum of selective expert sub-branches weighted by $\lambda$, training each expert individually. $\mathcal{L}_{a}$ is defined as:
\begin{equation}
  \begin{gathered}
  \mathcal{L}_{a}= \frac{1}{R}\sum(\Phi(\boldsymbol{\mathrm{f}}_0,\boldsymbol{\rho}_{\theta}) \mathrm{+} \Phi({\boldsymbol{\dot{y}}_0, \boldsymbol{y}_{0}^{ex}}) \mathrm{+} \sum\limits_{r=1}^{K}\lambda_r \Phi({\boldsymbol{\dot{y}}_r, \boldsymbol{y}_{r}^{ex}})) \\
  \lambda_r = \left\{
    \begin{aligned}
    1, \quad & \boldsymbol{\dot{y}}_r = \boldsymbol{\mathrm{f}}_0 \\
    0, \quad & \boldsymbol{\mathrm{otherwise}}
    \end{aligned}
   \right. 
  \label{eq:moe_loss}
  \end{gathered}
\end{equation}
Where  $\boldsymbol{\dot{y}}_r$ denotes the one-hot encoded label of expert index $r$, and $\boldsymbol{\mathrm{f}}_0$ represents the label of \emph{bag} $\boldsymbol{B}$, with \emph{R} as the mini-batch size. After optimizing the Dyn-MoE parameters, we train the denoising network $\mathcal{D}$ by minimizing the noise estimation loss $\mathcal{L}_{e}$ as in \cite{ho2020denoising}, defined as:
\begin{equation}
  \mathcal{L}_{e} = ||\epsilon - \epsilon_{\theta}(\boldsymbol{Z}, \boldsymbol{\mathrm{f}}_t, \boldsymbol{\rho}_{\theta}, \boldsymbol{t})||^2
  \label{eq:noise_loss}
\end{equation}
MExD enhances the vanilla diffusion model by conditioning each step’s estimation function on priors that incorporate knowledge from the \emph{bag}.

\begin{table*}[ht!]
  \centering
  \caption{\textbf{Quantitative Comparison of Our Results and State-of-the-Art Methods} on CAMELYON16, TCGA-NSCLC, and BRACS datasets. All models are retrained under their recommended settings for fairness, except for PAMIL~\cite{zheng2024dynamic}, whose results are directly cited. The best performance is in \textbf{bold}, and the second best is \underline{underlined}.}
  \vspace{-5pt}
  \resizebox{0.95\linewidth}{!}{
    \begin{tabular}{lccccccccc}
    \toprule
    {\multirow{2}[2]{*}{\textbf{Method}}} & \multicolumn{3}{c}{\textbf{Camelyon16}} & \multicolumn{3}{c}{\textbf{TCGA-NSCLC}} & \multicolumn{3}{c}{\textbf{BRACS}} \\
    \cmidrule(r){2-4} \cmidrule(r){5-7} \cmidrule(r){8-10}
    & {F1-score} & {ACC}   & {AUC}  
    & {F1-score} & {ACC}   & {AUC}   
    & {F1-score} & {ACC}   & {AUC} \\
    \midrule
    \rowcolor{gray!20}
    \multicolumn{10}{c}{CTransPath pre-trained with SRCL} \\
    \midrule
    ABMIL~\cite{ilse2018attention} & 89.81$\pm$0.69  & 90.70$\pm$0.63  & 92.41$\pm$0.78  & 89.67$\pm$0.81  & 89.71$\pm$0.80  & 95.87$\pm$0.57  &  62.54$\pm$1.16 & 68.29$\pm$0.86  & 80.44$\pm$0.80  \\
    DSMIL~\cite{li2021dual} & 88.84$\pm$0.73  & 89.73$\pm$0.74  & 92.05$\pm$1.17  & 90.65$\pm$0.73  & 90.67$\pm$0.72  & 96.40$\pm$0.53  & 69.73$\pm$1.82 & 72.82$\pm$1.06  & 86.70$\pm$0.43  \\
    CLAM-SB~\cite{lu2021data} &90.47$\pm$0.62 & 91.32$\pm$0.65 & 93.55$\pm$0.86 & 94.17$\pm$1.07 & 94.37$\pm$1.06  & 96.90$\pm$0.74  & 68.37$\pm$1.57 & 69.65$\pm$1.29 & 84.04$\pm$0.61 \\
    CLAM-MB~\cite{lu2021data} & 91.20$\pm$0.75 & 91.94$\pm$0.69 & 93.63$\pm$0.38 & 93.04$\pm$0.71 & 93.05$\pm$0.73 & 97.13$\pm$0.47 & 67.61$\pm$2.14 & 69.18$\pm$1.53 & 80.78$\pm$1.30\\
    TransMIL~\cite{shao2021transmil} & 93.69$\pm$0.56 & 94.19$\pm$0.44  & 95.03$\pm$0.67  & 90.94$\pm$0.88 & 90.95$\pm$0.89  & 95.89$\pm$0.23  & 70.17$\pm$1.80 & 72.00$\pm$1.29  & 85.18$\pm$0.56  \\
    DTFD-MIL~\cite{zhang2022dtfd} & 94.80$\pm$0.39  & 95.16$\pm$0.29  & 96.82$\pm$0.58  & 88.31$\pm$0.31  & 88.38$\pm$0.26  & 94.98$\pm$0.76  & 68.98$\pm$3.02 & 72.00$\pm$1.75  & 83.57$\pm$1.18  \\
    MHIM-MIL~\cite{tang2023multiple} & 94.67$\pm$0.68 & 94.99$\pm$0.76 & 96.14$\pm$0.55 & 91.34$\pm$0.81 & 91.68$\pm$0.49 & 96.73$\pm$0.65 & 72.41$\pm$1.63 & 73.26$\pm$0.96 & 84.79$\pm$0.82 \\
    MambaMIL~\cite{yang2024mambamil} & 93.98$\pm$0.71  & 94.42$\pm$0.65  & 96.06$\pm$0.78  & \underline{94.23$\pm$0.72}  & \underline{94.48$\pm$0.64}  &  97.16$\pm$0.40 &   68.37$\pm$0.84 & 70.59$\pm$0.83 & 83.39$\pm$1.01 \\
    ACMIL~\cite{zhang2023attention} & 93.80$\pm$0.32 & 93.99$\pm$0.38  & 94.85$\pm$0.72  & 94.08$\pm$1.01 & 94.10$\pm$0.98  & 96.80$\pm$0.76 & \underline{72.88$\pm$1.07} & \underline{74.35$\pm$0.52} & 85.70$\pm$0.73  \\
    IBMIL~\cite{lin2023IBMIL} & 95.02$\pm$0.65  & \underline{95.35$\pm$0.63} & 96.41$\pm$1.13  & 93.52$\pm$0.32  & 93.62$\pm$0.26  & \underline{97.45$\pm$0.53}  & 70.93$\pm$0.95 & 72.94$\pm$0.83 & \underline{85.84$\pm$0.71} \\
    WiKG~\cite{li2024dynamic} & 90.62$\pm$1.20 & 91.28$\pm$0.97  &92.73$\pm$1.82 & 93.23$\pm$0.53  & 93.24$\pm$0.52  & 95.95$\pm$1.24  & 69.00$\pm$0.60 & 71.06$\pm$0.64 & 84.19$\pm$0.59 \\
    $\mathrm{PAMIL}^{*}$~\cite{zheng2024dynamic} & \underline{96.10$\pm$0.60} & 94.70$\pm$0.90 & \underline{97.70$\pm$2.60} & {‡}  & {‡} & {‡}  & {‡}  & {‡} & {‡}   \\
    \rowcolor{c1}
    \emph{{\bf MExD}} & \textbf{97.29$\pm$0.41} & \textbf{97.48$\pm$0.37} & \textbf{98.87$\pm$0.43} & \textbf{96.51$\pm$0.29} & \textbf{96.53$\pm$0.23} & \textbf{98.13$\pm$0.17} & \textbf{75.17$\pm$0.42} & \textbf{76.13$\pm$0.58} & \textbf{88.08$\pm$0.26} \\
    \midrule
    \rowcolor{gray!20}
    \multicolumn{10}{c}{ViT pre-trained with MoCo V3} \\
    \midrule
    ABMIL~\cite{ilse2018attention} &85.04$\pm$1.06& 87.02$\pm$0.74  & 84.29$\pm$0.98  & 88.41$\pm$0.32 & 88.57$\pm$0.34  & 92.49$\pm$0.46  & 52.39$\pm$0.47 & 67.77$\pm$0.65  & 78.43$\pm$0.30  \\
    DSMIL~\cite{li2021dual} & 78.82$\pm$0.65  & 81.20$\pm$0.97  & 82.64$\pm$1.43  & 90.23$\pm$0.85  & 90.09$\pm$0.78  & 95.83$\pm$0.25  & 54.01$\pm$0.92 & 63.77$\pm$0.99  & 80.07$\pm$0.83  \\
    CLAM-SB~\cite{lu2021data} & 90.32$\pm$0.95 & 91.16$\pm$0.88 & 95.15$\pm$0.71  &94.38$\pm$0.53 & 94.38$\pm$0.52 & 96.93$\pm$0.23 &  69.67$\pm$1.54& 71.76$\pm$1.66 & 83.22$\pm$0.42 \\
    CLAM-MB~\cite{lu2021data} & 87.78$\pm$1.34 & 88.99$\pm$1.15  &  96.64$\pm$1.18 & 93.70$\pm$1.04 & 93.72$\pm$1.01 & 97.35$\pm$ 0.50 &  71.46$\pm$2.79 & 73.65$\pm$1.78 & 79.82$\pm$2.14 \\
    TransMIL~\cite{shao2021transmil} &93.33$\pm$0.60  & 93.80$\pm$0.63  & 95.05$\pm$0.44  & 93.51$\pm$0.71  & 93.62$\pm$0.72  & 96.40$\pm$0.41  & 74.04$\pm$0.31  & 74.82$\pm$0.64  & 87.15$\pm$0.39  \\
    DTFD-MIL~\cite{zhang2022dtfd} & 89.98$\pm$1.30 & 90.89$\pm$1.16  & 93.05$\pm$0.27  & 89.67$\pm$1.05  & 89.72$\pm$1.04  & 94.60$\pm$0.73  &  49.93$\pm$2.48 & 62.35$\pm$1.44  & 83.22$\pm$0.63  \\
    MHIM-MIL~\cite{tang2023multiple} & 93.24$\pm$1.40  & 93.70$\pm$0.77 & 97.20$\pm$0.60  & 90.48$\pm$1.50 & 92.17$\pm$0.81   & 94.95$\pm$0.53  & 62.76$\pm$1.31  & 71.28$\pm$1.26 & 86.20$\pm$0.69 \\
    MambaMIL~\cite{yang2024mambamil} & \underline{93.73$\pm$0.86} & \underline{94.11$\pm$0.88} & \underline{97.28$\pm$0.51} & 94.68$\pm$0.35 & 94.76$\pm$0.33  & \underline{97.76$\pm$0.36} &    61.63$\pm$2.48 & 69.88$\pm$1.05 & 85.62$\pm$0.84  \\
    ACMIL~\cite{zhang2023attention} & 86.69$\pm$0.72 & 87.75$\pm$0.65  & 89.44$\pm$0.42  & \underline{95.09$\pm$0.79}  & \underline{95.33$\pm$0.52} & 97.74$\pm$0.20 & 71.25$\pm$0.88 & 74.59$\pm$1.05  & 84.64$\pm$0.49  \\
    IBMIL~\cite{lin2023IBMIL} & 92.68$\pm$1.06 & 93.33$\pm$0.88  & 94.37$\pm$0.71  & 94.23$\pm$0.43 & 94.29$\pm$0.34 & 97.46$\pm$0.39  & \underline{74.74$\pm$0.61} & \underline{75.53}$\pm$0.98 & \underline{86.57$\pm$0.27} \\
    WiKG~\cite{li2024dynamic} & 86.28$\pm$0.96  &  87.60$\pm$0.55 & 90.06$\pm$1.45  & 92.73$\pm$0.87  & 92.76$\pm$0.85  &  97.52$\pm$0.44  & 63.99$\pm$2.15 & 66.83$\pm$1.54 & 82.52$\pm$0.88 \\
    \rowcolor{c1}
    \emph{{\bf MExD}} & \textbf{96.49$\pm$0.38} & \textbf{96.74$\pm$0.34} & \textbf{98.00$\pm$0.16} & \textbf{95.87$\pm$0.25} & \textbf{95.87$\pm$0.24} & \textbf{98.27$\pm$0.32} & \textbf{80.10$\pm$0.48} & \textbf{81.01$\pm$0.45} & \textbf{88.61$\pm$0.42} \\
    \bottomrule
    \end{tabular}%
    }
\vspace{-1em}
  \label{tab:tab1_com_sotas}%
\end{table*}%

\section{Experiments}
\label{sec:experiments}
\subsection{Experimental Setup} 

\noindent \textbf{Dataset and Evaluation Protocol.} We evaluate {MExD} on three public WSI classification benchmarks: Camelyon16~\cite{bejnordi2017diagnostic}, TCGA-NSCLC, and BRACS~\cite{brancati2022bracs}. Camelyon16 consists of H\&E-stained slides for breast cancer metastasis detection, while TCGA-NSCLC includes slides from two lung cancer subtypes: Lung Squamous Cell Carcinoma (LUSC) and Lung Adenocarcinoma (LUAD). BRACS covers three subtypes: Benign, Atypical, and Malignant. Following \cite{lin2023IBMIL}, the datasets are split into training and test sets: Camelyon16 (270/130), TCGA-NSCLC (836/210), and BRACS (395/87). As in \cite{li2021dual}, patches are extracted at $256\times256$ resolution with 20× magnification, yielding approximately 2.8M, 5.2M, and 2.4M patches for Camelyon16, TCGA-NSCLC, and BRACS, respectively. We conduct 5-fold cross-validation and report F1-score, Accuracy (ACC), and AUC with standard deviation (std) for robustness, while using PAvPU~\cite{mukhoti2018evaluating} to assess uncertainty.

\noindent \textbf{Training Details.} 
MExD is implemented in PyTorch and trained in two stages on an NVIDIA Tesla A100 GPU. First, we train the Dyn-MoE as an auxiliary model for 100 epochs using the RAdam optimizer~\cite{liu2020variance} with an initial learning rate of 2e-4 and weight decay of 1e-5. Next, Dyn-MoE is integrated into Diff-C to train the Diffusion classifier over 200 epochs using the Adam optimizer~\cite{kingma2014adam} with an initial learning rate of 1e-3. All learning rates are progressively reduced following a cosine schedule.



\subsection{Benchmarking on WSI Datasets}
\noindent {\bf{Quantitative Comparison with SOTAs.}} To evaluate MExD's performance, we compare it with 12 SOTA baselines on 3 widely-used benchmarks, including ABMIL~\cite{ilse2018attention}, DSMIL~\cite{li2021dual}, CLAM~\cite{lu2021data}, TransMIL~\cite{shao2021transmil}, DTFD-MIL~\cite{zhang2022dtfd}, MHIM-MIL~\cite{tang2023multiple}, MambaMIL~\cite{yang2024mambamil}, ACMIL~\cite{zhang2023attention}, IBMIL~\cite{lin2023IBMIL}, WiKG~\cite{li2024dynamic}, and PAMIL~\cite{zheng2024dynamic}. For fair comparison, we used the same instance-wise \emph{bags} encoded by ViT (pre-trained with MoCo V3~\cite{chen2021empirical}) and CTransPath (pre-trained with SRCL~\cite{wang2020visual}) and trained all models under their recommended settings.

\begin{table}[t!]
  \centering
  \renewcommand{\arraystretch}{1.0}
  \setlength{\tabcolsep}{3pt}
  \caption{\textbf{Quantitative Uncertainty Assessment} using PAvPU ($\alpha$-value=0.05)~\cite{mukhoti2018evaluating}, conducted via hypothesis testing with CTransPath~\cite{wang2020visual} as \(\boldsymbol{f}_{\mathbb{PFE}}\).}
  \vspace{-5pt}
    \resizebox{0.93\linewidth}{!}{
        \begin{tabular}{lcccccc}
        \toprule
       \makebox[0.18\linewidth][c]{\multirow{2}[2]{*}{{Method}}} & \multicolumn{2}{c}{Camelyon16} & \multicolumn{2}{c}{TCGA-NSCLC}& \multicolumn{2}{c}{BRACS}
       \\
       \cmidrule(r){2-3} \cmidrule(r){4-5} \cmidrule(r){6-7}
        & \makebox[0.13\linewidth][c]{ACC} & \makebox[0.13\linewidth][c]{PAvPU}
        & \makebox[0.13\linewidth][c]{ACC} & \makebox[0.13\linewidth][c]{PAvPU}
        & \makebox[0.13\linewidth][c]{ACC} & \makebox[0.13\linewidth][c]{PAvPU} \\
        \midrule        
        
        ACMIL\cite{zhang2023attention}       
                    & 94.57 & 93.80 & 94.29 & 93.81 & 74.12 & 76.47 \\ 

        IBMIL~\cite{lin2023IBMIL}       
                    & 94.57 & 93.02 & 93.81 & 94.29 & 74.12 & 75.29  \\ 
        
        TransMIL~\cite{shao2021transmil}    
                    & 94.57 & 94.57 & 91.90 & 91.90 & 72.94 & 72.94 \\ 
        
        MambaMIL~\cite{yang2024mambamil}    
                    & 95.35 & 95.35 & 94.76 & 94.76 & 71.76 & 71.76 \\ 
        \rowcolor{c1}
        \textbf{MExD} & \textbf{97.67} & \textbf{98.45} & \textbf{96.67} & \textbf{97.62} & \textbf{76.47} & \textbf{81.18} \\ 
        \bottomrule 
        \vspace{-3em}
   \end{tabular}%
    }
    \label{tab:pavpu}
\end{table}%

\begin{figure*}[t!]
  \centering
    \includegraphics[width=0.98\linewidth]{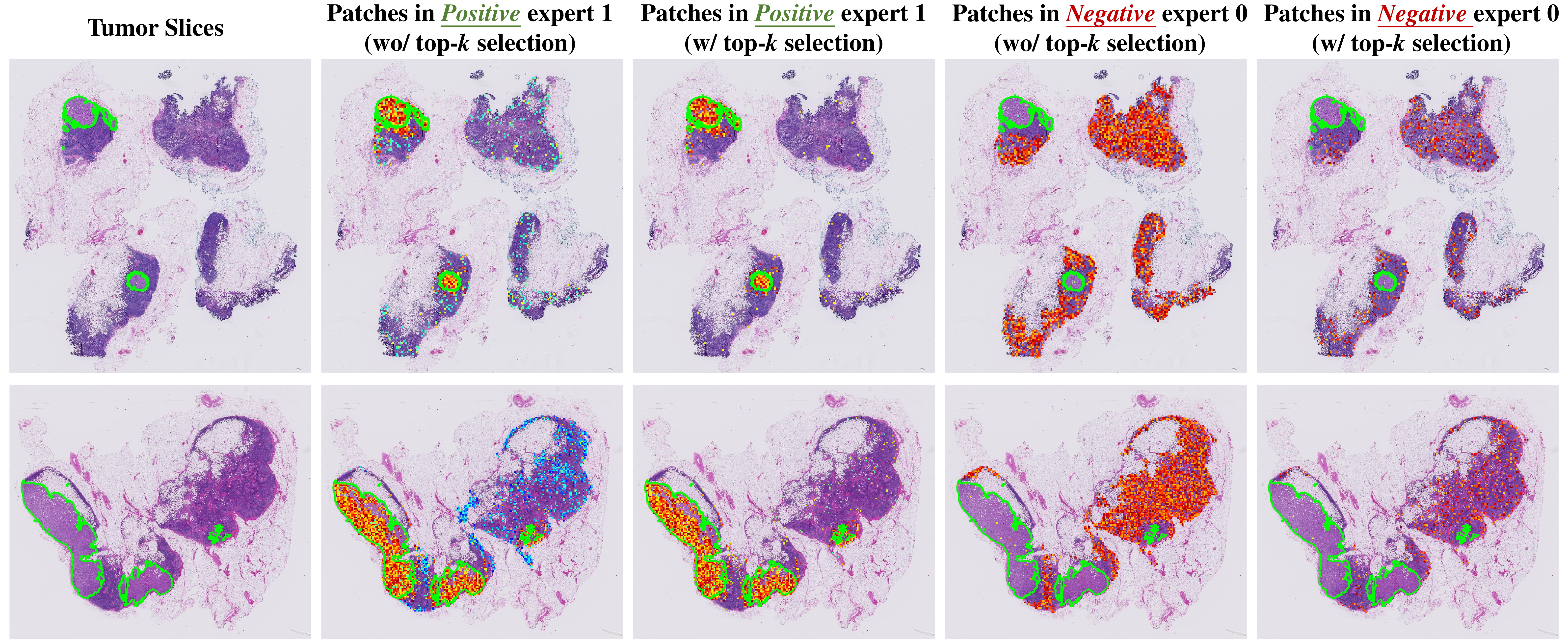}
    	\vspace{-5pt}
   \caption{\textbf{Distribution Visualization} of patch-wise router scores for \emph{positive} instances, where each score corresponds to a selected patch. In column 2, expert 1 effectively identifies \emph{positive} patches, then refines and retains the most representative ones (column 3) through score-based selection. Concurrently, expert 0 focuses on refining \emph{negative} instances. The {green edges} indicate cancerous region.}
   \label{fig:quantative_ana}
        \vspace{-1em}
\end{figure*}

As shown in Tab.~\ref{tab:tab1_com_sotas}, MExD achieves leading performance across all metrics and datasets, setting new benchmarks. MExD consistently enhances all feature extractors, indicating its agnosticism to instance features. Specifically, we analyze results by feature aggregator: \textbf{1) CTransPath.} On Camelyon16~\cite{bejnordi2017diagnostic}, MExD sets new benchmarks of 97.29\%, 97.48\%, and 98.87\% across all metrics, outperforming PAMIL~\cite{zheng2024dynamic} by 1.19\% in F1-score, 2.78\% in ACC, and 1.17\% in AUC, underscoring its strength in binary classification with unbalanced \emph{bags}. On TCGA-NSCLC, MExD improves by 2.31\%, 2.05\%, and 0.68\% across all metrics, achieving an ACC of 96.53\%, showing its capability in multi-class MIL. On BRACS~\cite{brancati2022bracs}, it surpasses ACMIL~\cite{zhang2023attention} by 2.29\% in F1-score, 1.78\% in ACC, and 2.38\% in AUC. \textbf{2) ViT.} MExD outperforms MambaMIL~\cite{yang2024mambamil} by 2.76\% in F1-score, 2.63\% in ACC, and 0.72\% in AUC on Camelyon16. On TCGA-NSCLC, we see an average increase of 0.68\% across metrics, while BRACS exhibits the largest gains of 5.36\%, 5.48\%, and 2.04\% in F1-score, ACC, and AUC, respectively. These results demonstrate MExD’s effectiveness in classifying diverse WSIs with tailored expert sub-branches and generative prediction reconstruction.

\noindent {\bf{Qualitative Interpretability.}} 
Fig.~\ref{fig:quantative_ana} illustrates Dyn-MoE’s ability to select key instances for WSI classification. The first column shows slides with cancerous regions outlined in green. In column 2, router scores from our \emph{positive} expert 1 are mapped onto the WSI, and patches without scores are excluded. Expert 1 identifies \emph{positive} patches effectively, despite some \emph{negative} interference, achieving instance sparsity. Notably, true \emph{positive} patches receive higher scores, while false \emph{positives} are scored lower. Column 3 demonstrates score-based refinement, where only higher-scoring patches are retained, yielding sparser instance sets controlled by sampling proportion $\alpha$. Concurrently, expert 0 filters redundant \emph{negative} instances.

\begin{table}[pt!]
  \centering
  \caption{\textbf{Performance Comparison Between MExD Variants} (using CTransPath~\cite{wang2020visual} as $\boldsymbol{f}_{\mathbb{PFE}}$), where $\Delta_{1}$, $\Delta_{2}$ represent Dyn-MoE and Diff-C, respectively.}
  	\vspace{-5pt}
    \resizebox{0.93\linewidth}{!}{
        \begin{tabular}{cccccccc}
        \toprule
        \makebox[0.06\textwidth][c]{\multirow{2}[2]{*}{{Method}}} & \multicolumn{3}{c}{TCGA-NSCLC} & \multicolumn{3}{c}{BRACS} & {\multirow{2}[2]{*}{{Avg.}}}\\
        \cmidrule(r){2-4} \cmidrule(r){5-7} 
        & \makebox[0.05\textwidth][c]{F1-score} & \makebox[0.05\textwidth][c]{ACC} & \makebox[0.05\textwidth][c]{AUC} & \makebox[0.05\textwidth][c]{F1-score} & \makebox[0.05\textwidth][c]{ACC} & \makebox[0.05\textwidth][c]{AUC} \\
        \midrule
        Base.                          & 91.37 & 91.43 & 96.00 & 68.59 & 70.59 & 84.80 & --\\
        +$\Delta_{1}$                  & 94.28 & 94.29 & 97.26 & 73.53 & 74.16 & 86.02 & --\\
        \rowcolor{c1}
        \multicolumn{1}{c}{Gap ($\%$)} & \textbf{+2.89} & \textbf{+2.86} & \textbf{+1.26} 
                                       & \textbf{+4.94} & \textbf{+3.57} & \textbf{+1.22} 
                                       & \textbf{+2.79} \\
        \midrule
        Base.                          & 91.37 & 91.43 & 96.00 & 68.59 & 70.59 & 84.80 & --\\
        +$\Delta_{2}$                  & 92.37 & 92.38 & 96.67 & 71.51 & 72.94 & 85.63 & --\\
        \rowcolor{c1}
        \multicolumn{1}{c}{Gap ($\%$)} & \textbf{+1.00} & \textbf{+0.95} & \textbf{+0.67}
                                       & \textbf{+2.92} & \textbf{+2.35} & \textbf{+0.83} 
                                       & \textbf{+1.45}\\
        \midrule
        Base.+$\Delta_{1}$             & 94.28 & 94.29 & 97.26 & 73.53 & 74.16 & 86.02 & --\\
        +$\Delta_{1}$+$\Delta_{2}$     & 96.66 & 96.67 & 98.29 & 75.95 & 76.47 & 88.14 & --\\
        \rowcolor{c1}
        \multicolumn{1}{c}{Gap ($\%$)} & \textbf{+2.38} & \textbf{+2.38} & \textbf{+1.03}
                                       & \textbf{+2.42} & \textbf{+2.31} & \textbf{+2.12}
                                       & \textbf{+2.11}\\
        \bottomrule 
        \vspace{-2.5em}
        
   \end{tabular}%
        }
\label{tab:equipment}%
\end{table}%

\begin{figure}[t!]
  \centering
    \includegraphics[width=0.9\linewidth]{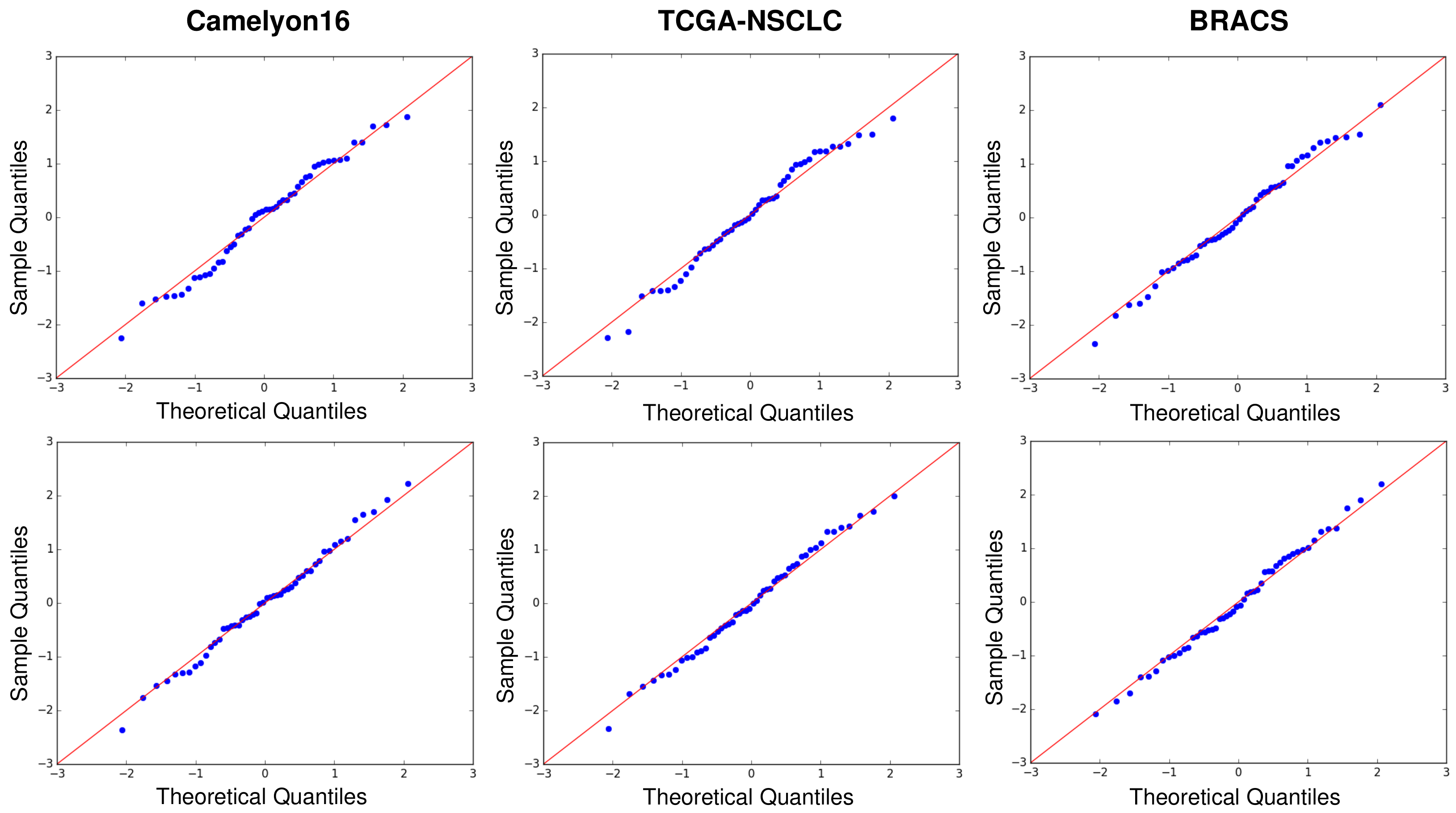}
    	\vspace{-5pt}
   \caption{\textbf{Normality Assumption Assessment:} Q-Q plots showing the probability difference between the top two predicted classes within a \emph{bag}, with two \emph{bags} selected from each benchmark for visualization.}
   \label{fig:qq_plot}
   \vspace{-20pt}
\end{figure}

\noindent {\bf{Hypothesis Testing based Uncertainty Estimation.}} 
To validate the enhanced confidence of our diffusion-based classifier in MExD, we employ a statistical testing approach for uncertainty estimation, inspired by \cite{fan2021contextual}. Hypothesis testing provides interpretable p-values. Specifically, we test whether the difference in probabilities of the top two predicted classes is statistically significant. 
For each \emph{bag}, we gather 100 posterior samples and conduct a paired two-sample t-test. If the null hypothesis (equal means) is rejected, it indicates that MExD is certain to this prediction.
Based on t-test results with $\alpha$-value = 0.05, we compute PAvPU~\cite{mukhoti2018evaluating}. As seen in Tab.~\ref{tab:pavpu}, MExD shows full certainty in correct predictions and uncertainty in some incorrect ones, outperforming other competitors. This enables safer human-AI collaboration~\cite{madras2018predict}. The t-test’s normality assumption is confirmed through Q-Q plots~\cite{ghasemi2012normality} (Fig.~\ref{fig:qq_plot}), as all points align closely with a 45-degree line. For a more comprehensive understanding and additional details, please refer to the {\bf Supplementary Materials}.

\subsection{Ablation Study}

\noindent {\bf{Ablation for Components in MExD}.}
We thoroughly investigate the impact of Dyn-MoE and Diff-C by testing MExD with various component configurations. Using a model with an Adapter and MLP classifier (MLP-C) as our baseline, we observe a 2.79\% average improvement across all metrics on TCGA-NSCLC and BRACS upon integrating Dyn-MoE (Tab.~\ref{tab:equipment}). Replacing MLP-C with Diff-C further boosts performance, showing a 1.45\% increase overall, with the F1-score for BRACS~\cite{brancati2022bracs} rising by 2.92\%. The fully configured MExD achieves optimal results, with a 4.90\% gain across metrics. Notably, applying Diff-C after Dyn-MoE yields an additional 3.45\% boost compared to its application after the baseline, highlighting Dyn-MoE's role in enhancing Diff-C's discriminative power.

\noindent {\bf{Compatibility for MIL Models.}} To evaluate the scalability of Dyn-MoE and Diff-C, we integrate them into four MIL methods (Sec~\ref{sec:mexd}) and test on TCGA-NSCLC. As shown in Tab.~\ref{tab:compatibility}, all baselines see notable gains. Unlike methods classifying all instances, Dyn-MoE’s inherent sparsity captures more representative instances, boosting accuracy. For instance, in MeanPooling~\cite{ilse2020deep}, it reduces redundant \emph{negatives}, improving F1-score by up to 14.04\%.

\begin{table}[t]
  \centering
  \renewcommand{\arraystretch}{1.2}
  \setlength{\tabcolsep}{3pt}
  \caption{\textbf{Compatibility Analysis of Dyn-MoE and Diff-C Across Four MIL Baselines} on TCGA-NSCLC (using ViT~\cite{chen2021empirical} as $\boldsymbol{f}_{\mathbb{PFE}}$), showing significant improvements across all baselines.}
  	\vspace{-5pt}
    \resizebox{0.9\linewidth}{!}{
        \begin{tabular}{l|ccc}
        
        \toprule
       \makebox[0.3\linewidth][c]{Method} & \makebox[0.23\linewidth][c]{F1-score} & \makebox[0.23\linewidth][c]{ACC} & \makebox[0.23\linewidth][c]{AUC} \\
        \midrule
        ABMIL~\cite{ilse2018attention}  & 89.96 & 90.00 & 93.29 \\ 
        \rowcolor{c1} MExD-ABMIL          & 93.80(\textbf{+3.84\%})  & 93.81(\textbf{+3.81\%}) & 96.17(\textbf{+2.88\%}) \\ 
        \midrule
        DSMIL~\cite{li2021dual}    & 91.42  & 91.43 & 95.96 \\ 
        \rowcolor{c1} MExD-DSMIL          & 94.74(\textbf{+3.32\%})  & 94.76(\textbf{+3.33\%}) & 98.13(\textbf{+2.17\%}) \\ 
        \midrule
        MeanPooling~\cite{ilse2020deep}   & 74.99  & 76.19 & 88.06 \\ 
        \rowcolor{c1} MExD-MeanPooling    & 89.03(\textbf{+14.04\%}) & 89.05(\textbf{+12.86\%}) & 94.35(\textbf{+6.29\%}) \\ 
        \midrule
        MaxPooling~\cite{ilse2020deep}    & 73.21 & 74.29 & 87.66  \\ 
        \rowcolor{c1} MExD-MaxPooling     & 86.17(\textbf{+12.96\%}) & 86.19(\textbf{+11.90\%}) & 93.58(\textbf{+5.92\%}) \\ 
        \bottomrule

   \end{tabular}%
        }
    \vspace{-12pt}
    \label{tab:compatibility}
\end{table}%

\noindent {\bf{Analysis for Sampling Ratio $\boldsymbol{\alpha}$.}}
We apply different sampling ratios, $\alpha_0$ for \emph{negative} and $\alpha_1$ for \emph{positive} instances, to address inherent imbalance. We systematically examine the influence of $\alpha_0$ and $\alpha_1$ on predictions, initially setting $\alpha_0$ to 0.5. As shown in Tab.~\ref{tab:sampling_ratio}, MExD’s performance improves with $\alpha_1$ adjusted from 0.75 to 0.5 but decreases as $\alpha_1$ drops to 0.1. Additionally, MExD continues to improve as $\alpha_0$ declines from 0.5 to 0.25. Our analysis indicates that setting $\alpha_0$ to half of $\alpha_1$ effectively mitigates imbalance, especially where \emph{negative} instances outnumber \emph{positive} ones. Furthermore, an $\alpha_0$ of 0.1 filters out too many relevant instances, reducing performance competitiveness.

\begin{table}[t!]
  \centering
  \renewcommand{\arraystretch}{1.0}
  \setlength{\tabcolsep}{3pt}
  \caption{\textbf{Sampling Ratio Analysis of $\boldsymbol{\alpha}$ in Dyn-MoE}. $\alpha_{0}$ and $\alpha_{1}$ are applied to the \emph{negative} and \emph{positive} expert sub-branches, respectively, using ViT~\cite{chen2021empirical} as $\boldsymbol{f}_{\mathbb{PFE}}$.}
  	\vspace{-5pt}
    \resizebox{0.9\linewidth}{!}{
        \begin{tabular}{cccccccc}
        
        \toprule
       \multicolumn{2}{c}{Settings} & \multicolumn{3}{c}{TCGA-NSCLC} & \multicolumn{3}{c}{BRACS}
       \\
       \cmidrule(r){1-2} \cmidrule(r){3-5} \cmidrule(r){6-8}
       \makebox[0.10\linewidth][c]{$\alpha_{0}$} & \makebox[0.10\linewidth][c]{$\alpha_{1}$} & \makebox[0.13\linewidth][c]{F1-score} & \makebox[0.13\linewidth][c]{ACC} & \makebox[0.13\linewidth][c]{AUC} & \makebox[0.13\linewidth][c]{F1-score} & \makebox[0.13\linewidth][c]{ACC} & \makebox[0.13\linewidth][c]{AUC}\\
        \midrule
        \multicolumn{1}{c|}{\multirow{4}[0]{*}{{0.50}}} 
        & \multicolumn{1}{c|}{0.75} & 94.28 & 94.29 & 96.84 & 77.77 & 78.82 & 87.60\\ 
        \multicolumn{1}{c|}{} 
        & \multicolumn{1}{c|}{0.50} & \cellcolor{c2}95.23 & \cellcolor{c2}95.24 & \cellcolor{c2}97.76 & \cellcolor{c2}78.81 & \cellcolor{c2}80.00 & \cellcolor{c2}88.00\\ 
        \multicolumn{1}{c|}{} 
        & \multicolumn{1}{c|}{0.25} & 94.76 & 94.76 & 96.80 & 77.88 & 78.82 & 87.86 \\ 
        \multicolumn{1}{c|}{} 
        & \multicolumn{1}{c|}{0.10} & 93.80 & 93.81 & 97.03 & 74.53 & 75.29 & 85.48 \\ 
        \midrule
        \multicolumn{1}{c|}{\multirow{4}[0]{*}{{0.25}}}
        & \multicolumn{1}{c|}{0.75} & 94.75 & 94.76 & 97.45 & 78.02 & 78.82 & 88.13\\ 
        \multicolumn{1}{c|}{} 
        & \multicolumn{1}{c|}{0.50} & \cellcolor{c1}{95.70} & \cellcolor{c1}95.71 & \cellcolor{c1}98.27 & \cellcolor{c1}80.33 & \cellcolor{c1}81.18 & \cellcolor{c1}88.72\\ 
        \multicolumn{1}{c|}{} 
        & \multicolumn{1}{c|}{0.25} & 94.76 & 94.76 & 97.93 & 77.02 & 77.65 & 87.88 \\ 
        \multicolumn{1}{c|}{} 
        & \multicolumn{1}{c|}{0.10} & 93.31 & 93.33 & 96.21 & 74.48 & 75.29 & 86.43\\ 
        \midrule
        \multicolumn{1}{c|}{\multirow{2}[0]{*}{{0.10}}}
        & \multicolumn{1}{c|}{0.75} & 92.85 & 92.86 & 92.40 & 68.99 & 72.94 & 84.87\\ 
        \multicolumn{1}{c|}{} 
        & \multicolumn{1}{c|}{0.50} & 90.89 & 90.95 & 90.78 & 71.01 & 71.76 & 83.23\\ 
        \bottomrule 
        
   \end{tabular}%
    }
    \vspace{-7pt}
    \label{tab:sampling_ratio}
\end{table}%

\noindent {\bf{Analysis for Prior Prediction $\boldsymbol{\rho}_{\theta}$}.}
MExD integrates prior prediction $\boldsymbol{\rho}_{\theta}$ as conditional expectations to reconstruct the input noise, improving the vanilla diffusion model for WSI classification. To highlight this modification’s effect, we set $\boldsymbol{\rho}_{\theta}=0$ as a control. As shown in Tab~\ref{tab:denoising_step}, introducing $\boldsymbol{\rho}_{\theta}$ boosts performance by 0.47\% on TCGA-DSCLC and 1.18\% on BRACS~\cite{brancati2022bracs}. Additionally, it reduces denoising steps T from 400 to 200, enhancing efficiency.

\begin{table}[t!]
  \centering
  \renewcommand{\arraystretch}{1.0}
  \setlength{\tabcolsep}{3pt}
  \caption{\textbf{Effectiveness of Prior Prediction $\boldsymbol{\rho}_{\theta}$ and Denoising Steps Ablation}, using ViT~\cite{chen2021empirical} as $\boldsymbol{f}_{\mathbb{PFE}}$, with T denoting the denoising steps.}
  	\vspace{-5pt}
    \resizebox{0.9\linewidth}{!}{
        \begin{tabular}{cccccccc}
        
        \toprule
       \makebox[0.18\linewidth][c]{\multirow{2}[2]{*}{{Method}}} & \multicolumn{3}{c}{TCGA-NSCLC}& \multicolumn{3}{c}{BRACS} & \multirow{2}[2]{*}{T}
       \\
       \cmidrule(r){2-4} \cmidrule(r){5-7}
        & \makebox[0.12\linewidth][c]{F1-score} & \makebox[0.12\linewidth][c]{ACC} & \makebox[0.12\linewidth][c]{AUC} & \makebox[0.12\linewidth][c]{F1-score} & \makebox[0.12\linewidth][c]{ACC} & \makebox[0.12\linewidth][c]{AUC} & \\
        \midrule
        \multicolumn{1}{c|}{\multirow{4}[0]{*}{\makecell[l]{MExD \\ w/o $\boldsymbol{\rho}_{\theta}$}}} 
        & 93.80 & 93.81 & 96.17 & 75.56 & 76.47 & 86.63 & 100\\ 
        
        \multicolumn{1}{c|}{} 
        & 94.28 & 94.29 & 96.88 & 77.74 & 78.82 & 86.76 & 200\\ 
        
        \multicolumn{1}{c|}{} 
        & 95.23 & 95.24 & 97.00 & 78.02 & 78.82 & 87.19 & 300\\ 
         
        \multicolumn{1}{c|}{} 
        & \cellcolor{c2}95.23 & \cellcolor{c2}95.24 & \cellcolor{c2}97.63 & \cellcolor{c2}79.17 & \cellcolor{c2}80.00 & \cellcolor{c2}87.72 & \cellcolor{c2}400\\ 
        \midrule
        \multicolumn{1}{c|}{\multirow{4}[0]{*}{\makecell[l]{MExD \\ w/ $\boldsymbol{\rho}_{\theta}$}}} 
        & 95.23 & 95.24 & 96.49 & 78.02 & 78.86 & 87.06 & 100\\ 
         
        \multicolumn{1}{c|}{} 
        & \cellcolor{c1}95.70 & \cellcolor{c1}95.71 & \cellcolor{c1}98.27 & \cellcolor{c1}80.15 & \cellcolor{c1}81.18 & \cellcolor{c1}88.72 & \cellcolor{c1}200\\ 
        
        \multicolumn{1}{c|}{} 
        & 95.70 & 95.71 & 98.17 & 80.33 & 81.18 & 87.79 & 300\\ 
         
        \multicolumn{1}{c|}{} 
        & 95.71 & 95.71 & 98.00 & 80.15 & 81.18 & 88.00 & 400\\ 
        
        \bottomrule 
        
   \end{tabular}%
    }
    \vspace{-10pt}
    \label{tab:denoising_step}
\end{table}%

\noindent {\bf{Analysis for Denoising Step T}.}
We investigate the effect of DDVM denoising steps T on prediction accuracy through ablation experiments with T from 100 to 400. As shown in Tab.~\ref{tab:denoising_step}, MExD improves steadily on TCGA-NSCLC and BRACS as T increases to 200, stabilizing beyond that. Without prior prediction $\rho_{\theta}$, performance drops, and T steps increase substantially to 400.

\section{Conclusion}
\label{sec: conclusion}
In this work, we introduce MExD, a novel WSI classification framework that combines a conditional generative approach with Mixture-of-Experts (MoE). MExD includes a Dyn-MoE aggregator that dynamically selects relevant instances, addressing data imbalance and improving representation. Its generative Diff-C also iteratively denoises Gaussian noise using expert insights and prior predictions, resulting in stable outcomes. This blend of MoE and generative classification offers a promising new approach to WSI classification with broader applications in computer vision.

\noindent {\bf Acknowledgement.} This research is supported by the National Research Foundation, Singapore under its AI Singapore Programme (AISG Award No: AISG2-TC-2021-003), and A*STAR Central Research Fund, the National Natural Science Foundation of China (62203089, 62303092, 62103084), and the Sichuan Science and Education Joint Fund General Project (No. 2024NSFSC2065).

\bibliographystyle{ieeenat_fullname}
\bibliography{main}






\end{document}